\documentclass[11pt,a4paper]{article}

\usepackage{multicol}
\usepackage{times}
\usepackage{latexsym}
\usepackage{url}
\usepackage{graphicx}
\usepackage{xcolor}
\usepackage{colortbl,booktabs}
\usepackage{tabu}
\usepackage{multirow} 
\usepackage{float}
\usepackage[hyperref]{acl2019}

\aclfinalcopy 

\title{ERNIE: Enhanced Representation through Knowledge Integration}
\author{
 Yu Sun,
 Shuohuan Wang,
 Yukun Li, 
 Shikun Feng \\
 \bfseries{
 Xuyi Chen,
 Han Zhang,
 Xin Tian,
 Danxiang Zhu,
 Hao Tian,
 Hua Wu}
 \\
 Baidu Inc. \\
 \{\texttt{sunyu02,wangshuohuan,liyukun01,fengshikun01,tianhao,wu\_hua}\}\texttt{@baidu.com} \\
}

\begin{document}
\bibliographystyle{acl_natbib}
\maketitle
\begin{abstract}
We present a novel language representation model enhanced by knowledge called ERNIE (Enhanced Representation through kNowledge IntEgration). 
Inspired by the masking strategy of BERT \cite{devlin2018bert}, ERNIE is designed to learn language representation enhanced by knowledge masking strategies, which includes entity-level masking and phrase-level masking. Entity-level strategy masks entities which are usually composed of multiple words.
Phrase-level strategy masks the whole phrase which is composed of several words standing together as a conceptual unit.
Experimental results show that ERNIE outperforms other baseline methods, achieving new state-of-the-art results on five Chinese natural language processing tasks including natural language inference, semantic similarity, named entity recognition, sentiment analysis and question answering. 
We also demonstrate that ERNIE has more powerful knowledge inference capacity on a cloze test.
\end{abstract}

\section{Introduction}

Language representation pre-training \cite{mikolov2013efficient, devlin2018bert} has been shown effective for improving many natural language processing tasks such as named entity recognition, sentiment analysis, and question answering. 
In order to get reliable word representation, neural language models are designed to learn word co-occurrence and then obtain word embedding with unsupervised learning.
The methods in Word2Vec \cite{mikolov2013efficient} and Glove \cite{pennington2014glove} represent words as vectors, where similar words have similar word representations. These word representations provide an initialization for the word vectors in other deep learning models. Recently, lots of works such as Cove \cite{mccann2017learned}, Elmo \cite{peters2018deep}, GPT \cite{radford2018improving} and BERT \cite{devlin2018bert} improved word representation via different strategies, which has been shown to be more effective for down-stream natural language processing tasks. 


The vast majority of these studies model the representations by predicting the missing word only through the contexts. These works do not consider the prior knowledge in the sentence.  
For example, In the sentence " \textit{Harry Potter is a series of fantasy novels written by J. K. Rowling}".\textit{ Harry Potter} is a novel name and  \textit{J. K. Rowling} is the writer. It is easy for the model  to predict the missing word of the entity \textit{ Harry Potter} by word collocations inside this entity without the help of long contexts. 
The model cannot predict \textit{ Harry Potter} according to the relationship between \textit{ Harry Potter} and \textit{J. K. Rowling}. 
It is intuitive that if the model learns more about prior knowledge, the model can obtain more reliable language representation.

In this paper, we propose a model called ERNIE (enhanced representation through knowledge integration) by using knowledge masking strategies. In addition to basic masking strategy, we use two kinds of knowledge strategies: phrase-level strategy and entity-level strategy. We take a phrase or a entity as one unit, which is usually composed of several words. All of the words in the same unit are masked during word representation training, instead of only one word or character being masked. In this way, the prior knowledge of phrases and entities are implicitly learned during the training procedure. 
Instead of adding the knowledge embedding directly, ERNIE implicitly learned the information about knowledge and longer semantic dependency, such as the relationship between entities, the property of a entity and the type of a event, to guide word embedding learning. This can make the model have better generalization and adaptability.


In order to reduce the training cost of the model, ERNIE is pre-trained on heterogeneous Chinese data, and then applied to 5 Chinese NLP tasks.
ERNIE advances the state-of-the-art results on all of these tasks. 
An additional experiment on the cloze test shows that ERNIE has better knowledge inference capacity over other strong baseline methods.

Our Contribution are as follows:

(1) We introduce a new learning processing of language model which masking the units such as phrases and entities in order to implicitly learn both syntactic and semantic information from these units.


(2) ERNIE significantly outperforms the previous state-of-the art methods on various Chinese natural language processing tasks.

(3) We released the codes of ERNIE and pre-trained models, which are available in  \url{https://github.com/PaddlePaddle/LARK/tree/develop/ERNIE} . 

\begin{figure*} 
\centerline{\includegraphics[width=375pt,height=200pt]{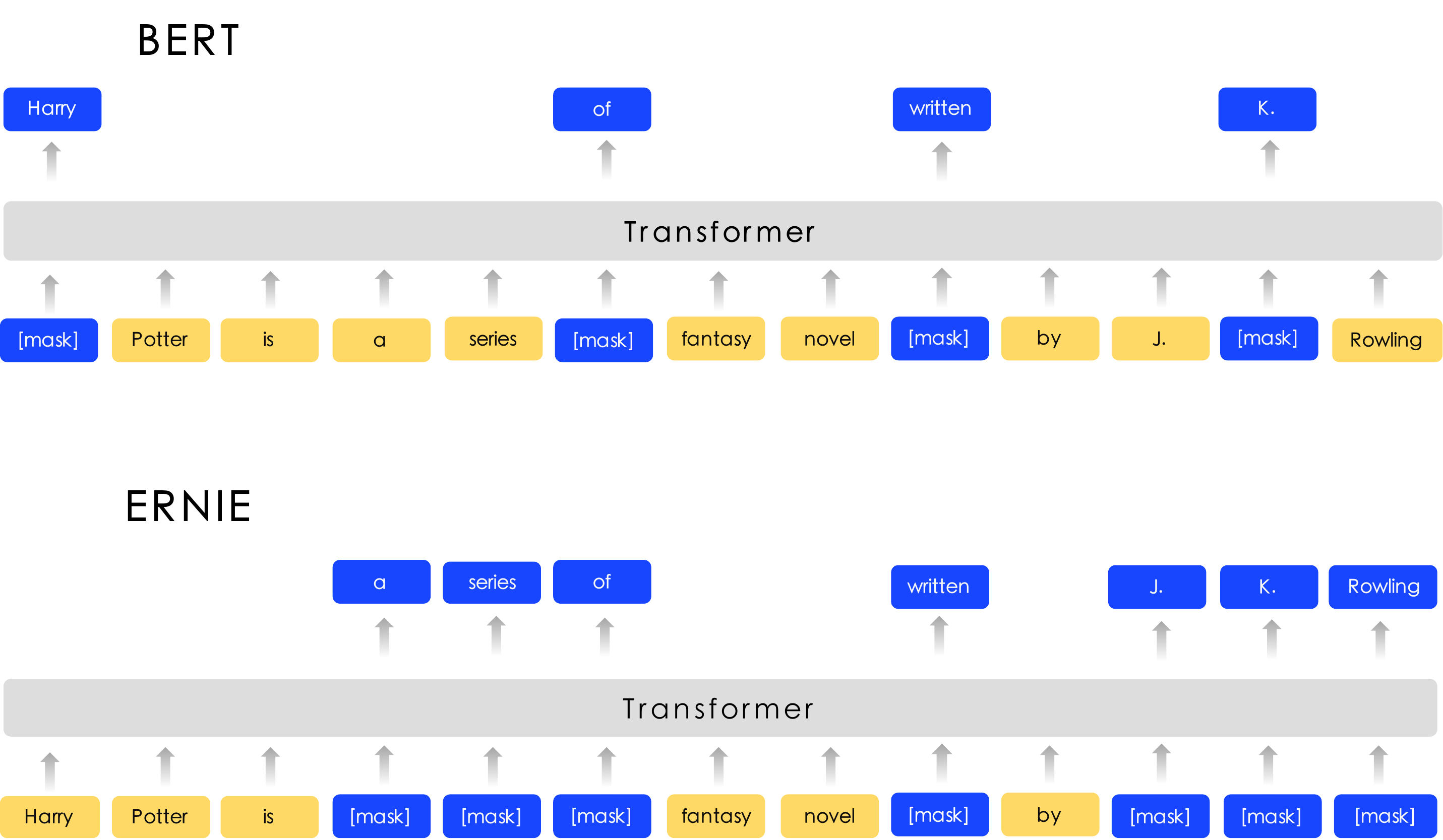}}
\caption{The different masking strategy between BERT and ERNIE}
\label{difference_bert_with_ernie}
\end{figure*}

\section{Related Work}
\subsection{Context-independent Representation}
Representation of words as continuous vectors has a long history. A very popular model architecture for estimating neural network language model (NNLM) was proposed in \cite{bengio2003neural}, where a feed forward neural network with a linear projection layer and a non-linear hidden layer was used to learn the word vector representation.

It is effective to learn general language representation by using a large number of unlabeled data to pretrain a language model. Traditional methods focused on context-independent word embedding. Methods such as Word2Vec \cite{mikolov2013efficient} and Glove \cite{pennington2014glove} take a large corpus of text as inputs and produces a word vectors, typically in several hundred dimensions. They generate a single word embedding representation for each word in the vocabulary.

\subsection{Context-aware Representation}
However, a word can have completely different senses or meanings in the contexts. Skip-thought \cite{kiros2015skip} proposed a approach for unsupervised learning of a generic, distributed sentence encoder. Cove \cite{mccann2017learned} show that adding these context vectors improves performance over using only unsupervised word and character vectors on a wide variety of common NLP tasks. ULMFit \cite{howard2018universal} proposed an effective transfer learning method that can be applied to any task in NLP. ELMo \cite{peters2018deep} generalizes traditional word embedding research along a different dimension. They propose to extract context-sensitive features from a language model. The GPT \cite{radford2018improving} enhanced the context-sensitive embedding by adapting the Transformer. 

BERT \cite{devlin2018bert} uses two different pretraining tasks for language modeling. BERT randomly masks a certain percentage of words in the sentences and learn to predict those masked words. Moreover, BERT learn to predict whether two sentences are adjacent. This task tries to model the relationship between two sentences which is not captured by traditional language models. Consequently, this particular pretraining scheme helps BERT to outperform state-of-the-art techniques by a large margin on various key NLP datasets such as GLUE \cite{wang2018glue} and SQUAD \cite{rajpurkar2016squad} and so on.

Some other researchers try to add more information based on these models. MT-DNN \cite{liu2019multi} combine pre-training learning and multi-task learning to improve the performances over several different tasks in GLUE \cite{wang2018glue}. GPT-2 \cite{radfordlanguage} adds task information into the pre-training process and adapt their model to zero-shot tasks. XLM \cite{lample2019cross} adds language embedding to the pre-training process which achieved better results in cross-lingual tasks.

\subsection{Heterogeneous Data}
Semantic encoder pre-trained on heterogeneous unsupervised data can improve the transfer learning performance. Universal sentence encoder \cite{google_use} adopts heterogeneous training data drawn from Wikipedia, web news, web QA pages and discussion forum. Sentence encoder \cite{google_smartreply} based on response prediction benefits from query-response pair data drawn from Reddit conversation. XLM \cite{lample2019cross} introduce parallel corpus to BERT, which is trained jointly with masked language model task. With transformer model pre-trained on heterogeneous data, XLM shows great performance gain on supervise/unsupervised MT task and classification task.

\section{Methods}

We introduce ERNIE and its detailed implementation in this section. We first describe the model's transformer encoder,and then introduce the knowledge integration method in Section 3.2. The comparisons between BERT and ERNIE are shown visually in Figure \ref{difference_bert_with_ernie}.

\begin{figure*} 
\centerline{\includegraphics[width=450pt]{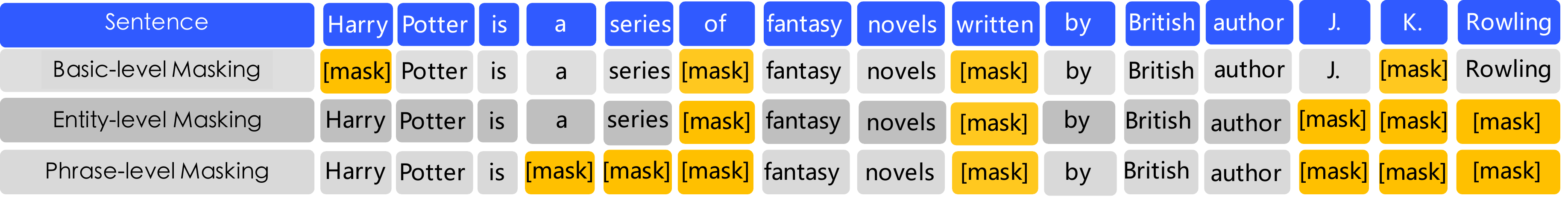}}
\caption{Different masking level of a sentence}
\label{Mixed-Strategy Masking}
\end{figure*}

\subsection{Transformer Encoder} 
ERNIE use multi-layer Transformer \cite{vaswani2017attention} as basic encoder like  previous pre-traning model such as GPT, BERT and XLM. 
The Transformer can capture the contextual information for each token in the sentence via self-attention, and generates a sequence of contextual embeddings. 

For Chinese corpus, we add spaces around every character in the CJK Unicode range and use the WordPiece \cite{wu2016google} to tokenize Chinese sentences. 
For a given token, its input representation is constructed by summing the corresponding token, segment and position embeddings. The first token of every sequence is the special classification embedding([CLS]).

\subsection{Knowledge Integration}
we use prior knowledge to enhance our pretrained language model. Instead of adding the knowledge embedding directly, we proposed a multi-stage knowledge masking strategy to integrate phrase and entity level knowledge into the Language representation. The different masking level of a sentence is described in Figure \ref{Mixed-Strategy Masking}.

\subsubsection{Basic-Level Masking}
The first learning stage is to use basic level masking, It treat a sentence as a sequence of basic Language unit, for English, the basic language unit is word, and for Chinese, the basic language unit is Chinese Character. In the training process, We randomly mask 15 percents of basic language units, and using other basic units in the sentence as inputs, and train a transformer to predict the mask units.
Based on basic level mask, we can obtain a basic word representation. Because it is trained on a random mask of basic semantic units, high level semantic knowledge is hard to be fully modeled.

\subsubsection{Phrase-Level Masking}
The second stage is to employ phrase-level masking. Phrase is a small group of words or characters together acting as a conceptual unit. For English, we use lexical analysis and chunking tools to get the boundary of phrases in the sentences, and use some language dependent segmentation tools to get the word/phrase information in other language such as Chinese. In phrase-level mask stage, we also use basic language units as training input, unlike random basic units mask, this time we randomly select a few phrases in the sentence, mask and predict all the basic units in the same phrase. At this stage, phrase information is encoded into the word embedding. 

\subsubsection{Entity-Level Masking}
The third stage is entity-level masking. Name entities contain persons, locations, organizations, products, etc., which can be denoted with a proper name. It can be abstract or have a physical existence. Usually entities contain important information in the sentences. As in the phrase masking stage, we first analyze the named entities in a sentence, and then mask and predict all slots in the entities. After three stage learning，a word representation enhanced by richer semantic information is obtained.

\begin{figure*} 
\centerline{\includegraphics[width=16cm]{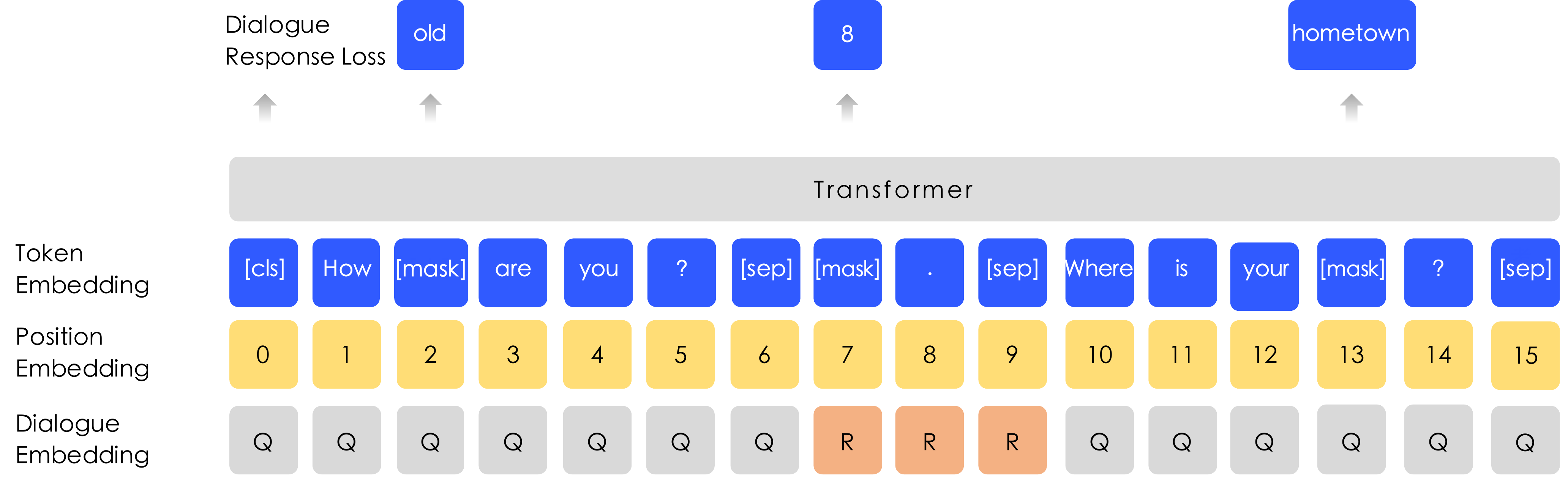}}
\caption{Dialogue Language Model. Source sentence: [cls] How [mask] are you ？ [sep] 8 . [sep] Where is your [mask] ? [sep]. Target sentence (words the predict): old, 8, hometown)
}
\label{DLM_fig}
\end{figure*}

\section{Experiments}
ERNIE was chosen to have the same model size as BERT-base for comparison purposes. ERNIE uses 12 encoder layers, 768 hidden units and 12 attention heads.
\subsection{Heterogeneous Corpus Pre-training}
ERNIE adopts Heterogeneous corpus for pre-training. Following \cite{google_use}, we draw the mixed corpus Chinese Wikepedia, Baidu Baike, Baidu news and Baidu Tieba. The number of sentences are 21M, 51M, 47M, 54M. respectively. Baidu Baike contains encyclopedia articles written in formal languages, which is used as a strong basis for language modeling. Baidu news provides the latest information about movie names, actor names, football team names, etc. Baidu Tieba is an open discussion forum like Reddits, where each post can be regarded as a dialogue thread. Tieba corpus is used in our DLM task, which will be discussed in the next section. 

 We perform traditional-to-simplified conversion on the Chinese characters, and upper-to-lower conversion on English letters. We use a shared vocabulary of 17,964 unicode characters for our model.

\subsection{DLM}
Dialogue data is important for semantic representation, since the corresponding query semantics of the same replies are often similar. ERNIE models the Query-Response dialogue structure on the DLM (Dialogue Language Model) task. As shown in figure \ref{DLM_fig}, our method introduces dialogue embedding to identify the roles in the dialogue, which is different from that of universal sentence encoder \cite{google_use}. ERNIE's Dialogue embedding plays the same roles as token type embedding in BERT, except that ERNIE can also represent multi-turn conversations (e.g. QRQ, QRR, QQR, where Q and R stands for "Query" and "Response" respectively). Like MLM in BERT, masks are applied to enforce the model to predict missing words conditioned on both query and response. What's more, we generate fake samples by replacing the query or the response with a randomly selected sentence. The model is designed to judge whether the multi-turn conversation is real or fake. 

The DLM task helps ERNIE to learn the implicit relationship in dialogues, which also enhances the model's ability to learn semantic representation. The model architecture of DLM task is compatible with that of the MLM task, thus it is pre-trained alternatively with the MLM task.

\subsection{Experiments on Chinese NLP Tasks}
ERNIE is applied to 5 Chinese NLP tasks, including natural language inference, semantic similarity, named entity recognition, sentiment analysis, and question answering.

\subsubsection{Natural Language Inference}
The Cross-lingual Natural Language Inference (XNLI) corpus \cite{liu2019multi} is a crowd-sourced collection for the MultiNLI corpus. The pairs are annotated with textual entailment and translated into 14 languages including Chinese. The labels contains contradiction, neutral and entailment. We follow the Chinese experiments in BERT\cite{devlin2018bert}.

\subsubsection{Semantic Similarity}
The Large-scale Chinese Question Matching Corpus (LCQMC) \cite{liu2018lcqmc} aims at identifying whether two sentences have the same intention. Each pair of sentences in the dataset is associated with a binary label indicating whether the two sentences share the same intention, and the task can be formalized as predicting a binary label.

\subsubsection{Name Entity Recognition}
The MSRA-NER dataset is designed for named entity recognition, which is published by Microsoft Research Asia. The entities contains several types including person name, place name, organization name and so on. This task can be seen as a sequence labeling task.

\subsubsection{Sentiment Analysis}
ChnSentiCorp \cite{song6chnsenticorp} is a dataset which aims at judging the sentiment of a sentence. It includes comments in several domains such as hotels, books and electronic computers. the goal of this task is to judge whether the sentence is positive or negative.

\subsubsection{Retrieval Question Answering}
The goal of NLPCC-DBQA dataset ( \url{http://tcci.ccf.org.cn/conference/2016/dldoc/evagline2.pdf}) is to select answers of the corresponding questions. The evaluation methods on this dataset include MRR \cite{voorhees2001overview} and F1 score.

\subsection{Experiment results}
The test results on 5 Chinese NLP tasks are presented in Table \ref{finetune_table}. It can be seen that ERNIE outperforms BERT on all tasks, creating new state-of-the-art results on these Chinese NLP tasks. For the XNLI, MSRA-NER, ChnSentiCorp and nlpcc-dbqa tasks, ERNIE obtains more than 1\% absolute accuracy improvement over BERT. The gain of ERNIE is attributed to its knowledge integration strategy. 


\begin{table*}[htbp]
  \caption{Results on 5 major Chinese NLP tasks}
  \label{finetune_table}
  \centering

\begin{tabular}{c|c|cl|cc}
\hline
\multirow{2}{*}{Task}       & \multirow{2}{*}{Metrics} & \multicolumn{2}{c|}{Bert}       & \multicolumn{2}{c}{ERNIE}                                        \\ \cline{3-6} 
                            &                          & dev                      & test & dev                             & test                            \\ \hline
XNLI                        & accuracy                 & 78.1                     & 77.2 & 79.9 (+1.8)                     & 78.4 (+1.2)                     \\
LCQMC                       & accuracy                 & 88.8                     & 87.0 & 89.7 (+0.9)                     & 87.4 (+0.4)                     \\
MSRA-NER                    & F1                       & 94.0                     & 92.6 & 95.0 (+1.0)                     & 93.8 (+1.2)                     \\
ChnSentiCorp                & accuracy                 & 94.6                     & 94.3 & 95.2 (+0.6)                     & 95.4 (+1.1)                     \\
\multirow{2}{*}{nlpcc-dbqa} & mrr                      & 94.7                     & 94.6 & 95.0 (+0.3)                     & 95.1 (+0.5)                     \\
                            & F1                       & \multicolumn{1}{l}{80.7} & 80.8 & \multicolumn{1}{l}{82.3 (+1.6)} & \multicolumn{1}{l}{82.7 (+1.9)}
\end{tabular}
\end{table*}

\begin{table*}[!htbp]
  \caption{XNLI performance with different masking strategy and dataset size}
  \label{ablation_tabel}
  \centering
  \begin{tabular}{llll}
    \toprule
    \cmidrule(r){1-2}
    pre-train dataset size & mask strategy & dev Accuracy& test Accuracy \\
    \midrule
    10\% of all & word-level(chinese character) & 77.7\% & 76.8\% \\
    10\% of all & word-level\&phrase-level & 78.3\% & 77.3\% \\
    10\% of all & word-level\&phrase-leve\&entity-level & 78.7\%& 77.6\% \\
    all & word-level\&phrase-level\&entity-level & 79.9 \%& 78.4\% \\
    \bottomrule
  \end{tabular}
\end{table*}

\begin{table*}[!htbp]
  \caption{XNLI finetuning performance with DLM}
  \label{DML_tabel}
  \centering\textbf{}
  
  \begin{tabular}{llll}
    \toprule
    \cmidrule(r){1-2}
    corpus proportion(10\% of all training data)  & dev Accuracy& test Accuracy \\
    \midrule
    Baike(100\%) & 76.5\% &75.9\% \\
    Baike(84\%) /  news(16\%) & 77.0\% & 75.8\% \\
    Baike(71.2\%)/  news(13\%)/ forum Dialogue(15.7\%) & 77.7\%& 76.8\% \\
    \bottomrule
  \end{tabular}
\end{table*}

\subsection{Ablation Studies}
To better understand ERNIE, we perform ablation experiments over every strategy of ERNIE in this section.

\subsubsection{Effect of Knowledge Masking Strategies}
We sample 10\% training data from the whole corpus to verify the effectiveness of the knowledge masking strategy. 
Results are presented in Table \ref{ablation_tabel}. We can see that 
adding phrase-level mask to the baseline word-level mask can improve the performance of the model. Based on this, we add the entity-level masking strategy，the performance of the model is further improved. 
In addition. The results also show that with 10 times larger size of the pre-training dataset, 0.8\% performance gain is achieved on XNLI test set.

\subsubsection{Effect of DLM}
Ablation study is also performed on the DLM task. 
we use 10\% of all training corpus with different proportions to illustrate the contributions of DLM task on XNLI develop set.
we pre-train ERNIE from scratch on these datasets, and report average result on XNLI task from 5 random restart of fine-tuning.
Detail experiment setting and develop set result is presented in Table \ref{DML_tabel}, We can see that 0.7\%/1.0\% of improvement in develop/test accuracy is achieved on this DLM task.

\begin{figure*} 
\centerline{\includegraphics[width=16cm]{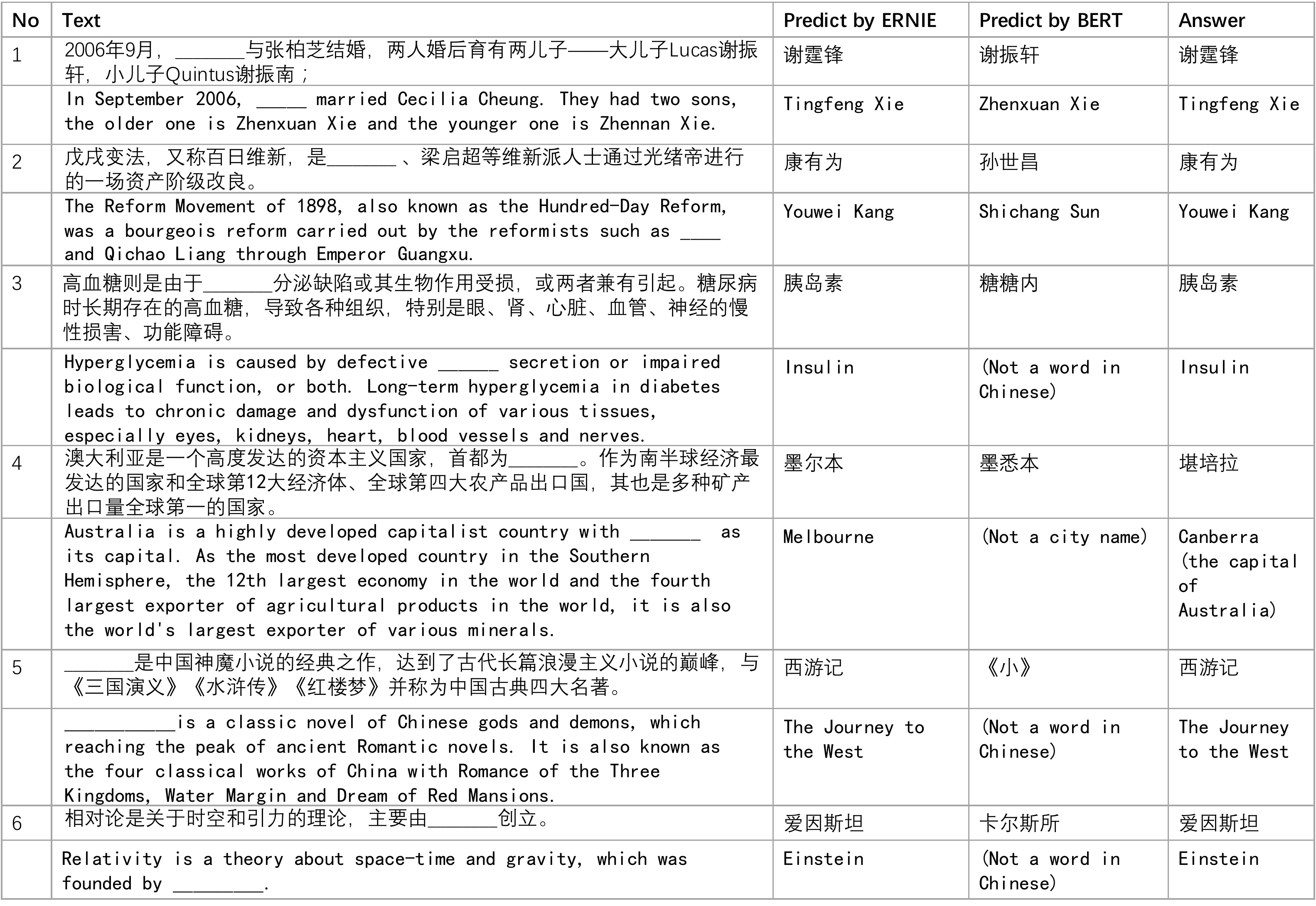}}
\caption{Cloze test}
\label{Cloze}

\end{figure*}

\subsection{Cloze Test}
To verify ERNIE's knowledge learning ability, We use several Cloze test samples \cite{taylor1953cloze} to examine the model. In the experiment, the name entity is removed from the paragraphs and the model need to infer what it is. Some cases are show in Figure \ref{Cloze}. We compared the predictions of BERT and ERNIE.

In case 1, BERT try to copy the name appeared in the context while ERNIE remembers the knowledge about relationship mentioned in the article. In cases 2 and Case 5, BERT can successfully learn the patterns according to the contexts, therefore correctly predicting the named entity type but failing to fill in the slot with the correct entity. on the contrary, ERNIE can fill in the slots with the correct entities. In cases 3, 4, 6, BERT fills in the slots with several characters related to sentences, but it is hard to predict the semantic concept. ERNIE predicts correct entities except case 4. Although ERNIE predicts the wrong entity in Case 4, it can correctly predict the semantic type and fills in the slot with one of an Australian city. In summary, these cases show that ERNIE performs better in context-based knowledge reasoning. 

\section{Conclusion}
In this paper, we presents a novel method to integrate knowledge into pre-training language model. Experiments on 5 Chinese language processing tasks show that our method outperforms BERT over all of these tasks. We also confirmed that both the knowledge integration and pre-training on heterogeneous data enable the model to obtain better language representation.

In future we will integrate other types of knowledge into semantic representation models, such as using syntactic parsing or weak supervised signals from other tasks. In addition We will also validate this idea in other languages.

\bibliography{acl2019}
\bibliographystyle{acl_natbib}

\end{document}